%% file: main.tex
\documentclass[letterpaper]{article} 
\usepackage[preprint]{aaai2027} 
\usepackage[hyphens]{url} 
\usepackage{graphicx} 
\usepackage{natbib} 
\usepackage{caption} 
\usepackage{algorithm}
\usepackage{algpseudocode}
\usepackage{newfloat}
\usepackage{listings}
\DeclareCaptionStyle{ruled}{labelfont=normalfont,labelsep=colon,strut=off}
\floatstyle{ruled}
\newfloat{listing}{tb}{lst}{}
\floatname{listing}{Listing}

\usepackage{amsmath}
\usepackage{amssymb}
\usepackage{booktabs}
\usepackage{xspace}
\usepackage{comment}

\usepackage{tabularx}
\usepackage{longtable}
\usepackage{array}
\usepackage{seqsplit}
\usepackage{tcolorbox}
\tcbuselibrary{breakable,skins}

\definecolor{PromptTitle}{HTML}{737373}
\definecolor{PromptBorder}{HTML}{B8B8B8}
\definecolor{PromptBody}{HTML}{F7F7F7}

\newtcolorbox{promptbox}[1]{
  enhanced,
  breakable,
  title={#1},
  colback=PromptBody,
  colframe=PromptBorder,
  colbacktitle=PromptTitle,
  coltitle=white,
  fonttitle=\sffamily\bfseries\small,
  fontupper=\small,
  boxrule=0.65pt,
  arc=1.8mm,
  outer arc=1.8mm,
  left=4.5mm,
  right=4.5mm,
  top=3.5mm,
  bottom=3.5mm,
  before skip=10pt,
  after skip=12pt,
  before upper={\setlength{\parindent}{0pt}\setlength{\parskip}{4pt}},
  segmentation style={dashed,PromptBorder,line width=0.6pt}
}

\newcommand{\promptpart}[1]{
  \par\noindent{\sffamily\bfseries #1}\par\smallskip
}
\newcommand{\manifesttask}[1]{
  \noindent\textcolor{black!55}{\raisebox{0.15ex}{\scriptsize$\bullet$}}\,
  \texttt{\seqsplit{#1}}\par
}

\newcommand{\manifestcount}[1]{
  {\normalfont\bfseries #1 template\ifnum#1=1\else s\fi}\par\smallskip
}

\newcommand{\tool}{CoAdapt-GUI\xspace}

\title{\tool: Joint Workflow Context and Policy Adaptation for Unseen GUI Applications}
\author{
Linqiang Guo\textsuperscript{\rm 1,2} \quad
Li Gu\textsuperscript{\rm 1,2} \quad
Zihuan Jiang\textsuperscript{\rm 3} \quad
Zhixiang Chi\textsuperscript{\rm 3} \quad
Siobhan Reid\textsuperscript{\rm 1,2} \\
Ziqiang Wang\textsuperscript{\rm 1,2} \quad
Yuanhao Yu\textsuperscript{\rm 4} \quad
Wei Liu\textsuperscript{\rm 1} \quad
Yang Wang\textsuperscript{\rm 1,2} \quad
Tse-Hsun (Peter) Chen\textsuperscript{\rm 1}
}
\affiliations{
\textsuperscript{\rm 1} Concordia University \quad
\textsuperscript{\rm 2} Mila -- Qu\'ebec AI Institute \quad
\textsuperscript{\rm 3} University of Toronto \quad
\textsuperscript{\rm 4} McMaster University
}

\begin{document}

\maketitle

\begin{abstract}
Mobile GUI agents remain brittle when deployed to applications absent from
source training. We study novel-app generalization under a limited target
interaction budget and without target demonstrations. We introduce \tool, a
test-time adaptation (TTA) framework that jointly adapts structured workflow
context and policy from the agent's own target-app rollouts and rewards. The
workflow context retains transferable procedures, failure modes, and
verification rules while excluding app-bound source details. This separation
allows reusable workflow knowledge to guide adaptation without transferring
source-interface state. For policy adaptation, task--context-matched
group-relative optimization updates a LoRA adapter on a frozen
vision--language model. Across two unseen-app evaluations, \tool reaches
$45.0\%$ on AndroidWorld-Generalization, compared with $37.5\%$ for the
reported Policy-Only TTA baseline, and raises AndroidWorld Plus performance
from $38.6\%$ to $\mathbf{52.9\%}$. These results show that
transfer-constrained workflow context provides substantial gains and that
joint policy adaptation further improves held-out performance.
\end{abstract}

\input{tex/intro}

\input{tex/related}

\input{tex/approach}

\input{tex/eval}

\input{tex/limitation}
\input{tex/conclusion}

\bibliography{aaai2027}

\input{appendix}


\end{document}

%% file: tex/intro.tex
\section{Introduction}

Mobile GUI agents execute natural-language instructions by interpreting application interfaces and carrying out multi-step actions. Recent advances in GUI grounding, trajectory synthesis, and reinforcement learning have substantially improved their performance on established benchmarks~\citep{cheng2024seeclick,hong2024cogagent,openmobile2026,mobilerl2025}. However, these evaluations generally assume a predefined collection of applications and workflows~\citep{rawles2024androidworld,xu2024androidlab,mobileworld2025}. In deployment, an agent may instead encounter an application whose interface and task procedures were absent from training. Adapting to such applications from limited interaction is therefore essential for effective operation beyond the original training environment.

This challenge is particularly pronounced under cross-application shifts. AndroidWorld-Generalization distinguishes generalization to unseen task instances, templates, and applications. Source-side online RL improves a 7B policy by 26.1 percentage points on unseen instances, but the gain drops to 8.3 points on unseen applications~\citep{gu2026generalization}. Success on new tasks within a familiar interface therefore does not imply effective behavior in an unfamiliar application.

An unseen application can expose both interface-specific and procedural gaps. An agent may understand the goal but fail to ground actions to unfamiliar interface elements, or it may execute individual actions correctly yet lack the workflow and completion conditions needed to finish the task. Updating the policy can change how the agent interprets and acts on the new interface, while adapting an external workflow context can preserve procedures learned from target-side experience. Updating only one state may therefore leave an important source of failure unresolved.

Existing work only partially addresses this target-side adaptation problem. AndroidWorld-Generalization updates the policy from the agent's own target-app rollouts while leaving its workflow context fixed~\citep{gu2026generalization}. UI-Mem jointly learns experience memory and policy during source-side training, then transfers the resulting knowledge to unseen applications~\citep{xiao2026uimem}. This leaves open whether an agent can use interactions collected after encountering a new application to adapt both its workflow context and policy before evaluation on held-out target tasks.

We introduce \textbf{\tool}, a test-time adaptation (TTA) framework that maintains and jointly updates these two states from target-app rollouts and task-level rewards. Its context channel contrasts successful and failed traces to revise transferable procedures, failure patterns, and completion checks while excluding app-bound details. Its policy channel updates a lightweight LoRA adapter through a task--context-conditioned group-relative objective~\citep{shao2024deepseekmath}, while keeping the VLM backbone frozen. Both updates are derived from the same rollout groups within each adaptation round, and the resulting context and policy are frozen before evaluation on held-out target tasks.

We evaluate \tool under two levels of unseen-app generalization. 
AndroidWorld-Generalization~\citep{gu2026generalization} evaluates new
instances of task types encountered during target adaptation. In this setting,
\tool reaches 45.0\%, exceeding the reported Policy-Only TTA baseline of 37.5\% by 7.5 percentage points. To test whether adaptation
transfers beyond the task types encountered during target interaction, we
construct AndroidWorld Plus with disjoint adaptation and evaluation task types within each target app. Here, Context-Only TTA improves the Base Policy from 38.6\% to 48.1\%, while \tool further raises success to
\textbf{52.9\%}. These results show that target-grounded workflow adaptation provides substantial gains and that jointly adapting the policy achieves the best overall performance in both settings.

This paper makes three contributions:
\begin{itemize}
    \item \textbf{Joint target-side adaptation from autonomous interaction.}
    We introduce a test-time adaptation framework that updates separate
    workflow-context and policy states using only the agent's own target-app
    rollouts and task rewards, without target demonstrations or access to
    held-out evaluation signals.

    \item \textbf{Transfer-constrained context--policy adaptation.}
    We separate transferable workflow knowledge from app-bound source state
    and coordinate two reward-guided updates from the same target interactions.
    Validated reward differences revise the workflow context, while policy
    credit is computed only among rollouts sharing the same task and context
    condition.

    \item \textbf{Evaluation across instance- and task-type generalization.}
    We evaluate \tool on the released AndroidWorld-Generalization unseen-app
    split and construct AndroidWorld Plus to hold out entire task types during
    target adaptation. \tool reaches 45.0\% and 52.9\% in the two settings,
    respectively, achieving the best overall result in both settings.
\end{itemize}

%% file: tex/related.tex
\section{Related Work}
\label{sec:related}

\subsection{Mobile GUI Agents and Generalization}

Mobile GUI agents commonly combine a vision--language policy with an interaction loop for action execution and task verification. GUI-specific pretraining and high-resolution perception improve visual grounding and action prediction~\citep{cheng2024seeclick,hong2024cogagent,gou2024uground}, while OpenMobile and MobileRL scale supervised or online policy learning from executable trajectories~\citep{openmobile2026,mobilerl2025}. UI-Mem also evaluates zero-shot transfer to held-out applications by retrieving hierarchical memory accumulated during source-side policy training~\citep{xiao2026uimem}. These approaches acquire transferable behavior on source applications before evaluation; we instead study how an agent can continue adapting after encountering an unseen application.

AndroidWorld, AndroidLab, and MobileWorld provide programmatically evaluated
environments for studying GUI agents across apps and multi-app
workflows~\citep{rawles2024androidworld,xu2024androidlab,mobileworld2025}.
Related studies examine generalization across websites, tasks, applications,
and app categories~\citep{deng2023mind2web,li2024androidcontrol}. Most closely,
AndroidWorld-Generalization separates unseen instances, templates, and apps,
and shows that policy adaptation on target-app interactions can improve
unseen-app performance~\citep{gu2026generalization}. We use its released
setting for comparison, but adapt both workflow context and the policy from
autonomously collected target rollouts.

\subsection{Test-Time Context and Policy Adaptation}

Test-time adaptation updates a deployed model using data from its target environment~\citep{sun2020ttt,wang2021tent}. Recent methods extend this idea to language-model reasoning and interactive agents~\citep{zuo2025ttrl,zweiger2025seal,chen2026ttaagents}. Few-shot GUI methods such as LearnAct and AdaptAgent also use target experience, but rely on demonstrations rather than the agent's own reward-bearing interaction~\citep{liu2025learnact,verma2024adaptagent}. Our setting instead uses executable rewards from autonomously collected trajectories and evaluates adaptation on held-out target tasks.

External memory provides a complementary in-context adaptation channel.
AppAgent records explored app functionality, Agent Workflow Memory abstracts
reusable routines, and Mobile-Agent-E and MobiMem accumulate experience as
evolving notes or prompt memory~\citep{yang2023appagent,wang2024awm,wang2025mobileagente,liu2025mobimem}. These methods demonstrate the value of explicit workflow knowledge while
generally keeping the underlying policy fixed. Unlike UI-Mem, which varies
memory guidance within each policy-optimization group to internalize guided
behavior~\citep{xiao2026uimem}, \tool treats context and policy as separate target-side states. Reward differences across context variants revise the workflow state, while policy credit is computed only among rollouts sharing the same task and context condition. Agent-SAMA additionally represents app execution as a finite-state machine (FSM) for planning and recovery~\citep{guo2025agentsama}. In contrast, \tool uses FSM-grounded context to constrain which workflow knowledge may transfer across applications, excluding app-bound source state from transfer.

E-SPL jointly optimizes a global free-text prompt and a policy from shared
rollouts~\citep{zhang2026espl}. \tool instead combines validated evolution of
structured workflow context with reward-based LoRA adaptation. Their overlap is
therefore limited to the high-level idea of joint context--policy adaptation.

%% file: tex/approach.tex
\section{\tool: Test-Time Context--Policy Adaptation}
\label{sec:method}

We study test-time adaptation to an unseen GUI application. 
Our central premise is that test-time behavior depends on two complementary adaptive states. An explicit workflow state $M_t$ provides contextual guidance about procedures, failure conditions, recovery strategies, and completion checks. A parametric state $\theta_t$ controls the
policy's underlying visual--action behavior. Rollouts in the target application provide executable feedback for adapting both states. 

Co-adapting these states introduces two forms of interference. First, source experience may mix portable workflow knowledge with application-specific screens, identifiers, geometry, and navigation paths, causing negative transfer on an unseen interface. Meanwhile, changes to the workflow context alter the effective policy input, so reward differences across context conditions need
not reflect differences in policy behavior.

\begin{figure*}[t]
    \centering
    \includegraphics[width=0.8\textwidth]{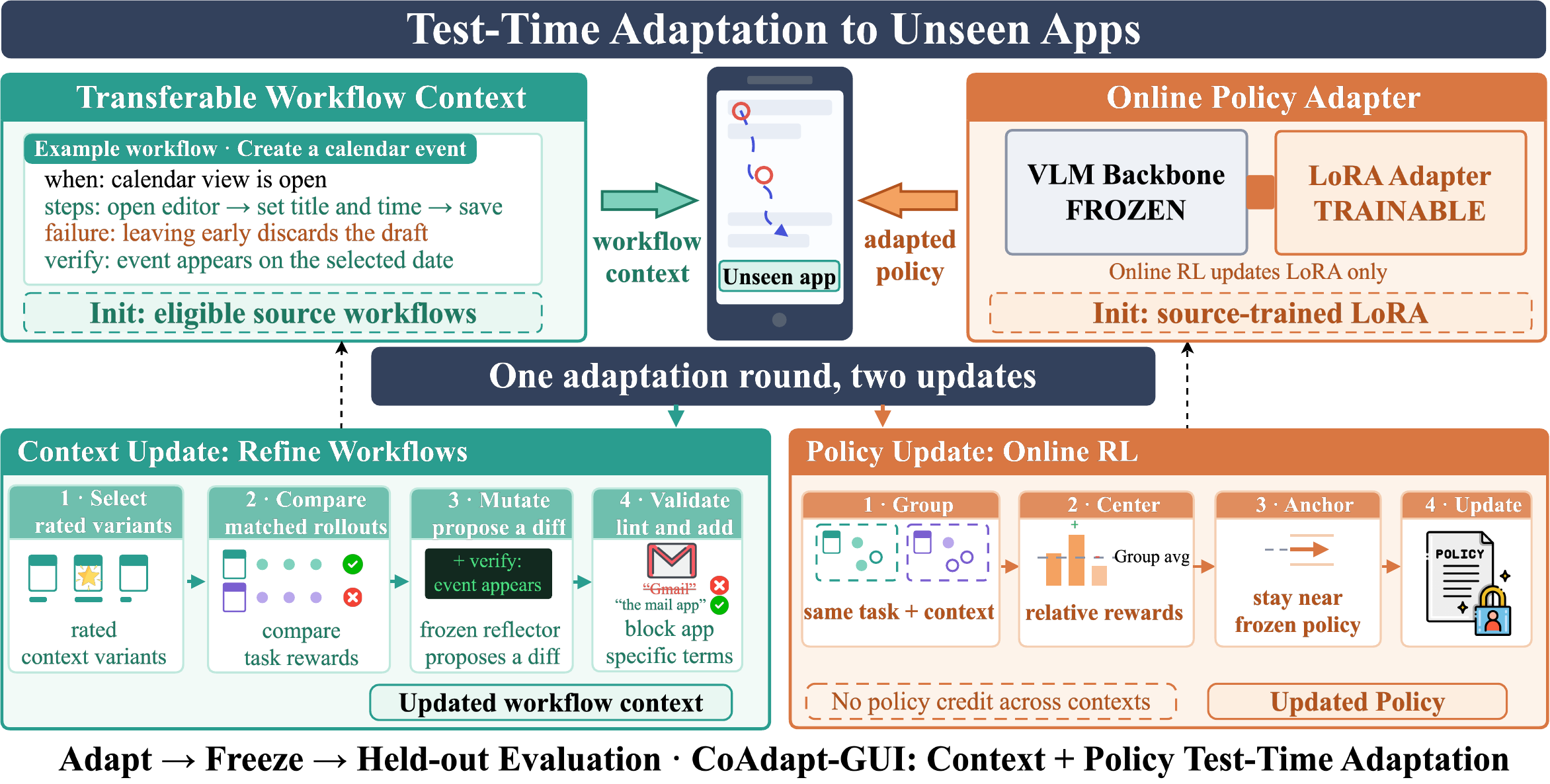}
    \caption{\tool uses matched target-app rollouts to refine transferable workflow context and update a lightweight policy adapter on a frozen VLM. Context-Only TTA executes only the context branch, whereas \tool executes both branches using the same rollout batches. The adapted context and policy are frozen before held-out evaluation.}
    \label{fig:evofsm-overview}
\end{figure*}

We propose \tool, which combines transfer-constrained workflow adaptation with
task--context-matched policy learning. The context channel prevents app-bound source details from crossing application boundaries and stores target-grounded revisions separately. The policy channel computes relative advantages only among rollouts sharing the same task and context condition. The updates are coupled through interaction: the current workflow state shapes the trajectories used for policy learning, while the current policy generates the successes and failures used to revise workflow knowledge. 
Figure~\ref{fig:evofsm-overview} summarizes the resulting adaptation loop.

\subsection{Test-Time Adaptation Problem}
\label{sec:method-setup}

Let $\mathcal{A}_{\mathrm{src}}$ and $\mathcal{A}_{\mathrm{tgt}}$ denote the
source and target application sets. In the strict unseen-app setting,
$\mathcal{A}_{\mathrm{src}}\cap\mathcal{A}_{\mathrm{tgt}}=\varnothing$.
More generally, target trajectories never enter source initialization, and application-specific interaction knowledge from one application is not transferred to another.

For test-time adaptation, tasks are partitioned into disjoint adaptation and
evaluation sets,
$\mathcal{Q}^{\mathrm{adapt}}$ and
$\mathcal{Q}^{\mathrm{eval}}$.
Only tasks in $\mathcal{Q}^{\mathrm{adapt}}$ may update the deployed agent;
evaluation tasks and outcomes remain unavailable throughout adaptation.

The workflow state for task $q$ combines an immutable source workflow context
$M_0(q)$ with a target-grounded state $M_t^{\mathrm{tgt}}$:
\[
M_t(q)
=
\left(
M_0(q),M_t^{\mathrm{tgt}}
\right),
\quad
C_t(q)
=
\operatorname{Render}\!\left(M_t(q),q\right).
\]
At interaction step $k$, the effective policy conditions on the task, current
observation $o_k$, interaction history $h_k$, and rendered workflow context:
\[
u_k
\sim
\pi_{\theta_t}
\left(
\cdot
\mid
q,o_k,h_k,C_t(q)
\right).
\]
Executing the action sequence produces a trajectory \(\tau=\{(o_k,u_k)\}_{k=1}^{K}\) and a task reward \(r(\tau)\).

The parametric state $\theta_t$ contains the trainable LoRA parameters of an otherwise frozen vision--language policy. Together with the target environment, $M_t(q)$ and $\theta_t$ induce the rollout distribution
\[
\tau
\sim
d\!\left(
\cdot
\mid
q,M_t(q),\theta_t
\right).
\]
Updating either state therefore changes the interaction data available for
updating the other.

Given an interaction budget $B$, adaptation returns final states
$(M^\star,\theta^\star)$. Both are frozen before evaluation, and their
generalization is estimated only on
$\mathcal{Q}^{\mathrm{eval}}$. Adaptation rewards are therefore training
signals rather than the final evaluation objective.

\subsection{Transfer-Constrained Workflow Context}
\label{sec:method-context}

Source trajectories mix reusable procedural knowledge with application-specific interaction details. 
Procedures, failure conditions, and completion checks may remain useful across applications with similar functionality, whereas screen layouts, visible strings, resource identifiers,
coordinates, and source-specific navigation paths may cause negative transfer.

We represent each transferable workflow entry as
\[
w
=
\langle c,P,F,V\rangle,
\]
where $c$ specifies when the workflow applies, $P$ describes an abstract
procedure, $F$ records failure or recovery conditions, and $V$ specifies
observable or executable completion checks. These fields are stored in a typed
schema so that applicability, execution guidance, failure handling, and
verification can be validated and revised separately.

For each source application $a$, we construct a finite-state machine (FSM) grounded workflow context with two components: an app-bound state $M_{\mathrm{app}}^{a}$ and a transferable state $M_{\mathrm{tr}}^{a}$. The app-bound state is instantiated as a screen-transition FSM recording concrete screens, action-conditioned transitions, visible interface cues, and resource-level information. This representation makes the expected effect of each action explicit, supporting the identification of failed transitions and possible recovery paths~\citep{guo2025agentsama}.

The transferable state contains workflow entries of the form defined above, abstracted from the same source trajectories used to construct the screen-transition FSM. Rather than encoding source-specific screens or
transitions, it describes what should be accomplished, which failures should be avoided, and how task completion should be verified. 

An eligibility predicate,
$\operatorname{Eligible}_{\mathrm{tr}}(w)\in\{0,1\}$,
determines which entries may cross application boundaries. A schema validator and linter reject entries containing app names, package or resource identifiers, concrete widget labels, coordinates, task-instance values, and other app-bound content. Eligibility does not imply that an entry will help the target; it excludes identifiable source-specific information, while target rollout rewards subsequently determine utility.

A frozen synthesizer constructs the app-bound and transferable source
states from audited trajectories. Let $\mathcal{W}_{\mathrm{src}}$ denote the
set of workflow entries $w$ such that $w\in M_{\mathrm{tr}}^{a}$ for some
source application $a\in\mathcal{A}_{\mathrm{src}}$ and
$\operatorname{Eligible}_{\mathrm{tr}}(w)=1$. These entries are grouped by
functional category and consolidated into the source library
\[
\mathcal{L}_{\mathrm{src}}
=
\operatorname{Consolidate}
\left(
\mathcal{W}_{\mathrm{src}}
\right).
\]

For target task $q$, initialization retrieves bounded, eligibility-constrained protocol-permitted source workflow context,
$M_0(q)=\operatorname{Retrieve}(\mathcal{L}_{\mathrm{src}},q)$.
Strict unseen-app evaluation excludes all app-bound source states. If the target
belongs to a functional category absent from the source pool, retrieval returns
an empty prior rather than forcing unrelated guidance. 

The source library remains fixed during target adaptation. Target interactions
instead maintain the separate target-grounded state
$M_t^{\mathrm{tgt}}$. Rendering selects task-relevant entries from the source workflow context and target-grounded state while preserving their provenance. This
separation prevents target adaptation from overwriting reusable source
knowledge while allowing target experience to refine or supplement the guidance
used during execution.

\subsection{Joint Context and Policy Adaptation}
\label{sec:method-adaptation}

Target interactions can update both the workflow state and the policy
parameters. At adaptation round $t$, the controller selects a task
$q_t\in\mathcal{Q}^{\mathrm{adapt}}$ and one or more already materialized
context variants from the current population. It then collects matched
rollouts under controlled reset conditions:
\[
D_t
=
\left\{
(\tau_j,r_j,\kappa_j,z_j)
\right\}_{j=1}^{N_t}.
\]
Here, \(\kappa_j\) identifies the context variant that generated trajectory \(\tau_j\), and \(z_j\) records its reset identity. Only context variants present before collecting \(D_t\) can receive rating updates from their rewards. The trajectories are also added to the policy buffer $\mathcal{B}$ with their task and context identities.

The two channels reuse the same rollout stream but operate at different
frequencies. Each matched batch updates the ratings of the evaluated context variants and may produce a validated child for subsequent evaluation. The LoRA parameters are updated only when \(\mathcal{B}\) contains valid task--context comparison groups. The updates are therefore interaction-coupled rather than jointly differentiable: the current context shapes the trajectories used for policy learning, while the current policy determines the successful and failed behaviors available for future context revisions.

\paragraph{Reward-guided context update.}
For the repeated seeded tasks used in our evaluation, the controller maintains a population of TrueSkill-rated context variants~\citep{herbrich2007trueskill}. At each round, it samples already materialized variants, evaluates them on matched tasks and resets seeds, and updates their ratings using the resulting task rewards. A frozen reflector then contrasts successful and failed traces from the evaluated variants and proposes a typed workflow revision to a high-rated parent.

The resulting child enters the population only after passing schema,
provenance, and lint checks. Because the child is proposed after collecting the current round's rollouts, it does not inherit their rewards. Its utility is determined only when it is sampled and evaluated in subsequent matched rollouts. At the end of adaptation, the highest-rated validated workflow state is selected for deployment.

\paragraph{Task--context-matched policy update.}
The policy update learns online from trajectories generated by the current policy rather than from a fixed dataset. Because
the context condition changes the effective policy input, trajectories are
partitioned into groups sharing both task and context:
\[
G(q,\kappa)
=
\left\{
j
\;\middle|\;
q_j=q,\;
\kappa_j=\kappa
\right\}.
\]
For a group $G$ of size $n_G$, we compute
\[
\bar r_G
=
\frac{1}{n_G}
\sum_{j\in G}r_j,
\qquad
A_j
=
\frac{r_j-\bar r_G}{s_G},
\]
where \(s_G\) is the normalization factor: it is set to one for mean-centered advantages, or to the within-group standard deviation plus a small constant for standardized advantages. Singleton and constant-reward groups provide no active policy gradient.

For trajectory $\tau_j$, we average log-probabilities over its
action-generation units:
\[
\ell_j(\theta)
=
\frac{1}{|\tau_j|}
\sum_{m\in\tau_j}
\log
\pi_{\theta}
\left(
u_{j,m}\mid x_{j,m}
\right),
\]
where $x_{j,m}$ contains the task, observation, history, and rendered context
available when action unit $u_{j,m}$ is generated. Let \(\mathcal{B}_{\mathrm{act}}\subseteq\mathcal{B}\) contain trajectories belonging to nondegenerate groups. The resulting online group-relative objective~\citep{shao2024deepseekmath} is
\[
\mathcal{L}_{\mathrm{policy}}
=
-
\frac{1}{|\mathcal{B}_{\mathrm{act}}|}
\sum_{j\in\mathcal{B}_{\mathrm{act}}}
A_j\ell_j(\theta)
+
\beta
\mathcal{R}
\left(
\theta;\pi_{\mathrm{anchor}}
\right),
\]
where \(\mathcal{R}\) denotes the frozen-policy regularizer when anchoring is enabled. The vision--language backbone remains frozen, and only the LoRA parameters are optimized~\citep{hu2021lora}. Restricting comparisons to $G(q,\kappa)$ controls variation due to task difficulty and contextual guidance when computing
relative advantages. The policy buffer $\mathcal{B}$ stores only trajectories collected since the most recent LoRA update. The policy remains fixed while these trajectories are collected. Once the update threshold is reached, at most one LoRA update is attempted, and the buffer is cleared, regardless of whether it contains an
active comparison group. No trajectory is reused after the policy changes. Each trajectory retains the task and rendered-context identity under which it was collected.

Algorithm~\ref{alg:adaptation} summarizes the complete schedule. Each matched comparison evaluates only already materialized context variants; revisions proposed from its traces enter the population for subsequent evaluation. LoRA
is updated only when the policy buffer contains nondegenerate task--context groups.
\begin{algorithm}[t]
\caption{CoAdapt-GUI test-time context--policy adaptation}
\label{alg:adaptation}
\begin{algorithmic}[1]
\Require Source workflow library $\mathcal{L}_{\mathrm{src}}$,
policy initialization $\theta_0$, adaptation tasks
$\mathcal{Q}^{\mathrm{adapt}}$, frozen reflector, interaction budget $B$
\Ensure Frozen workflow state $M^\star$ and policy parameters $\theta^\star$

\State Retrieve the protocol-permitted source workflow context $M_0$
\State Initialize target-grounded state
$M^{\mathrm{tgt}}\gets\varnothing$
\State Initialize policy $\theta\gets\theta_0$ and buffer
$\mathcal{B}\gets\varnothing$

\While{interaction budget remains}
    \State Select a task $q\in\mathcal{Q}^{\mathrm{adapt}}$
    \State Select already materialized context variants for $q$
    \State Collect matched rollouts $D$ and record their task, context,
    and reset identities
    \State Update the ratings of the evaluated variants using rewards in $D$
    \State Propose and validate a child revision from traces in $D$
    \If{the child passes all validation checks}
        \State Add the child to the population for subsequent evaluation
    \EndIf
    \State Add the same trajectories in $D$ to $\mathcal{B}$
    \If{LoRA adaptation is enabled and $\mathcal{B}$ is ready}
    \State Update LoRA from nondegenerate groups in $\mathcal{B}$, if any
    \State Clear $\mathcal{B}$
    \EndIf
\EndWhile

\State $M^\star\gets$ highest-rated validated workflow state
\State $\theta^\star\gets\theta$
\State Freeze $(M^\star,\theta^\star)$
\State \Return $(M^\star,\theta^\star)$
\end{algorithmic}
\end{algorithm}

\subsection{Source Initialization and Frozen Evaluation}
\label{sec:method-deployment}
 
During source preparation, the frozen synthesizer constructs the app-bound and transferable source states. Eligible workflow entries are consolidated into $\mathcal{L}_{\mathrm{src}}$. Separately, a shared LoRA adapter is trained on source-app trajectories while workflow context is held fixed. The resulting
context and policy initializations are independently usable: either may be deployed without enabling adaptation of the other.

At test time, initialization follows the evaluation protocol. Strict unseen-app evaluation excludes all app-bound source interaction knowledge, and a target from a novel functional
category begins without a source workflow prior. The policy is initialized from a source-side checkpoint selected using source metadata. Target trajectories and held-out outcomes never influence source construction or checkpoint routing.

Algorithm~\ref{alg:adaptation} then uses only
$\mathcal{Q}^{\mathrm{adapt}}$. The source library remains immutable, while target interactions update the separate target-grounded workflow state and, when enabled, the LoRA parameters. Once the interaction budget is exhausted, the selected workflow state and final policy adapter are frozen. No task, trajectory, or reward from $\mathcal{Q}^{\mathrm{eval}}$ may alter either state.

%% file: tex/eval.tex
\section{Experiments}
\label{sec:exp}

We evaluate \tool in two unseen-app settings. \textbf{(1)~AndroidWorld-Generalization} follows the released setup and tests adaptation to new instances of task templates encountered during adaptation, enabling direct
comparison with the reported Policy-Only TTA
baseline~\citep{gu2026generalization}. \textbf{(2)~AndroidWorld Plus}, an experimental extension of AndroidWorld~\citep{rawles2024androidworld}, separates
adaptation and evaluation by task template to test transfer to new task types within unseen apps.

\subsection{Experimental Setup}
\label{sec:exp-protocol}
Each experiment setting separates source data, target adaptation tasks, and held-out target evaluation tasks. Source data is used to obtain the initial policy and reusable workflow context. During target adaptation, the agent may update its workflow context, policy, or both, depending on the configuration. All updates stop before evaluation; no held-out task, trajectory, reward, or outcome is used for adaptation or checkpoint selection. A \emph{task template} defines a reusable
task type, whereas a \emph{task instance} supplies its concrete arguments. Our primary metric is success rate (SR), defined as the percentage of held-out evaluation episodes successfully completed according to the benchmark's programmatic evaluator.

\paragraph{Compared methods.} 

We mainly compare five configurations in each setting. \textbf{The Base Policy} uses the initial policy without workflow context or a target-side update. \textbf{Static Context Transfer} supplies source workflows as frozen context. \textbf{Policy-Only TTA} keeps the workflow context fixed and updates only the policy adapter from target adaptation rollouts. \textbf{Context-Only TTA} keeps the policy fixed and builds additional workflow context from target adaptation rollouts. 
\textbf{\tool} also updates a LoRA policy adapter while freezing the VLM backbone. These configurations isolate source workflow transfer, target-side context adaptation, policy adaptation, and their joint use.
Context-Only TTA and \tool receive the same adaptation tasks and rollout budget, but are run independently and can therefore collect different policy-dependent trajectories; their difference reflects the complete joint procedure, not a controlled estimate of the LoRA update alone. Experiments were run on NVIDIA H200 GPUs with 141~GB of memory; one 20-round target-app adaptation run required approximately 9--10 GPU-hours on average.

\subsection{New Task Instances in Unseen Apps}
\label{sec:exp-awg}

\paragraph{Setup.}
We follow the released unseen-app setting of AndroidWorld-Generalization~\citep{gu2026generalization}. Its source split
contains 12 apps, 62 task templates, and 905 training instances. All
configurations start from the released step-500 UI-TARS-7B checkpoint~\citep{qin2025uitarspioneeringautomatedgui}, trained on this split. The target split contains five disjoint apps, with eight adaptation instances per app (40 total) and 48 held-out evaluation instances. Adaptation and evaluation cover the same 16 templates but use instances generated with non-overlapping seeds, thereby testing transfer to new instances of task types seen during adaptation.

We use the released manifests throughout. All adaptive methods use the same eight adaptation instances and nominal 50-step schedule per target app, and all methods are evaluated on the same 48 held-out instances. The Base Policy and Policy-Only TTA values are taken directly from AndroidWorld-Generalization~\citep{gu2026generalization}.

\begin{table*}[t]
\footnotesize
\centering
\setlength{\tabcolsep}{3pt}
\begin{tabular*}{\textwidth}{@{\extracolsep{\fill}}llccccc@{}}
\hline
\textbf{Method} & \textbf{Config.} & \textbf{Adapt.} &
\textbf{Steps/app} & \textbf{Eval.} & \textbf{Success Rate (\%)} &
\textbf{Source} \\
\hline
The Base Policy~\citep{gu2026generalization}
& no TTA & 0 & 0 & 48 & 27.10 & Reported \\
Static Context Transfer
& static context & 0 & 0 & 48 & $28.75\pm2.28$ & Ours (5 runs) \\
Policy-Only TTA~\citep{gu2026generalization} 
& policy only & 40 & 50/app & 48 & 37.50 & Reported \\
Context-Only TTA
& context only & 40 & 50/app & 48 & $35.00\pm1.74$ & Ours (5 runs) \\
\textbf{\tool}
& context + policy & 40 & 50/app & 48 & $\mathbf{45.00\pm1.86}$ & Ours (5 runs) \\
\hline
\end{tabular*}
\caption{\textbf{Setting 1---New Task Instances in Unseen Apps (AndroidWorld-Generalization).} Success rates are measured on 48 held-out instances of task templates used during target adaptation. Reported results are from AndroidWorld-Generalization~\citep{gu2026generalization}; our results are mean $\pm$ standard deviation over five runs.}
\label{tab:aw-gu}
\end{table*}

\paragraph{Results.}
The reported Base Policy and Policy-Only TTA achieve 27.10\% and 37.50\%, respectively. Our \tool configuration performs best at $\mathbf{45.00\%\,\pm\,1.86}$, 7.5 points above the reported Policy-Only TTA result. Context-Only TTA reaches \(35.00\%\,\pm\,1.74\), while Static Context Transfer reaches \(28.75\%\,\pm\,2.28\). Under the same adaptation instances and rollout budget, \tool outperforms Context-Only TTA by 10.00 points, indicating that co-adapting workflow context and the policy is more effective than adapting context alone in this setting. 

\subsection{New Task Templates in Unseen Apps}
\label{sec:exp-awplus}

\paragraph{Setup.} To evaluate transfer to task templates not observed during adaptation, we construct AndroidWorld Plus by extending AndroidWorld~\citep{rawles2024androidworld} with three apps from B-MoCA~\citep{lee2024bmoca} and three from AndroidLab~\citep{xu2024androidlab}. After executable-task filtering, the benchmark contains 25 apps and 191 task templates. We assign 12 apps with 96 templates to the source set and the remaining 13 apps with 95 templates to the disjoint target set. 

All AndroidWorld Plus configurations use Qwen3-VL-8B-Instruct~\citep{qwen3vl2025} with a frozen backbone and start from a LoRA adapter trained only on the 12 source apps. Checkpoint selection occurs before target adaptation without using target rollouts or rewards, and each matched comparison uses the same initialization. Policy-Only TTA updates only the LoRA adapter while keeping its initial source
context fixed. Supplementary Section~1 provides the filtering procedure and complete app-level manifest.

Within each target app, adaptation and evaluation use disjoint task templates. Across the target apps, 60 templates yield a pool of 300 adaptation instances, while 35 held-out templates yield 105 evaluation episodes using non-overlapping seeds. Each adaptive configuration runs for 20 rounds per app with up to two context variants per round. Context-Only TTA, Policy-Only TTA, and \tool collect four rollouts per selected task--context condition ($\leq160$ per app). Context-Only TTA uses its batches only for workflow adaptation, Policy-Only TTA uses them only for LoRA adaptation, and
\tool uses them for both. Supplementary Section~3 provides the remaining adaptation details. This setting therefore tests whether experience from some task types transfers to unseen task types within the same target apps.



\paragraph{Source-category coverage.}
Following Android Control~\citep{li2024androidcontrol}, we use Google Play categories as a coarse, externally defined boundary for source-workflow retrieval. Source workflows are eligible only when their app category matches that of the target app. Under this protocol, six target apps are \textbf{Category-Shared} and can retrieve source workflows, whereas seven are
\textbf{Category-Novel}, for which retrieval returns no source workflow. Static Context Transfer therefore supplies no workflow context for Category-Novel Apps. Context-Only TTA and \tool instead initialize an empty target-grounded workflow state, construct workflow entries from target adaptation rollouts and rewards, and refine them in subsequent rounds. This
breakdown assesses whether target-side adaptation remains effective without category-matched source experience. Supplementary Section~1 provides the complete app allocation and implementation details.

\begin{table*}[t]
\footnotesize
\centering
\setlength{\tabcolsep}{3.5pt}
\begin{tabular*}{\textwidth}{@{\extracolsep{\fill}}llccccc@{}}
\hline
\textbf{Method}
& \textbf{Config.}
& \textbf{Adapt.}
& \textbf{Steps/app}
& \textbf{Category-Shared Apps}
& \textbf{Category-Novel Apps}
& \textbf{Overall SR} \\
\hline
The Base Policy
& no TTA
& 0
& 0
& 47.2\%
& 29.4\%
& 38.6\% \\

Static Context Transfer
& static context
& 0
& 0
& 56.5\%
& 29.4\%
& 43.3\% \\

Policy-Only TTA
& policy only
& 300
& 20/app
& 53.7\%
& 25.5\%
& 40.0\% \\

Context-Only TTA
& context only
& 300
& 20/app
& 63.9\%
& 31.4\%
& 48.1\% \\

\textbf{\tool{}}
& context + policy
& 300
& 20/app
& \textbf{70.4\%}
& \textbf{34.3\%}
& \textbf{52.9\%} \\
\hline
\end{tabular*}
\caption{\textbf{Setting 2---New Task Templates in Unseen Apps (AndroidWorld Plus).} Success rates are measured over 105 held-out episodes whose task templates are disjoint from target adaptation. Category-Shared and Category-Novel group episodes by whether the target app's category is represented in the source pool.}
\label{tab:aw-plus}
\end{table*}

\paragraph{Results.}
Our \tool performs best, raising overall success from 38.6\% to 52.9\%, a gain of 14.3 points. The cumulative context path---Static Context Transfer followed by Context-Only TTA---reaches 48.1\%, 9.5 points above the Base Policy, whereas Policy-Only TTA reaches 40.0\%, a gain of only 1.4 points. \tool is a further 4.8 points above Context-Only TTA. This pattern is consistent with evolved workflow guidance producing more informative target experience for policy learning.

On Category-Shared Apps, \tool reaches 70.4\%, compared with 53.7\% for Policy-Only TTA and 63.9\% for Context-Only TTA. Category-Novel Apps provide a built-in control for source transfer. Because retrieval is empty in this group, Static Context Transfer exactly matches the Base Policy at 29.4\%. Policy-Only TTA falls to 25.5\%, whereas Context-Only TTA and \tool improve success to 31.4\% and 34.3\%. 

Separately, a representative Chrome adaptation trace illustrates how context evolution can affect the policy-learning signal. All evaluated variants received zero task reward during the first 11 rounds. In round 12, an evolved variant reached a mean task reward of 0.25 over four rollouts, while the root context remained at zero. Context evolution thus exposed a task-success signal that was not observed under the original workflow guidance.

%% file: tex/limitation.tex
\section{Limitations}
\label{sec:limitations}

\paragraph{Evaluation scope and leakage control.}
Target-side adaptation can overestimate generalization through information leakage or evaluation on overly similar target tasks. We prevent leakage by using disjoint source and target app pools, restricting all updates and model selection to the adaptation split, and freezing both adapted states before held-out evaluation. We further evaluate two levels of task separation: AndroidWorld-Generalization uses disjoint instances generated from the same templates, whereas AndroidWorld Plus uses disjoint adaptation and evaluation templates within every target app. Fixed manifests, repeated runs, and a common evaluation harness are used across matched configurations.

\paragraph{Transfer and context reliability.}
Source workflows may contain interface-specific assumptions or errors
introduced during reflection, causing negative transfer to a new application. \tool addresses this risk in two stages. Before transfer, it separates app-bound screen-transition state from reusable workflow knowledge and applies eligibility, schema, provenance, and lint checks before a source workflow enters the transferable library. During target adaptation, a proposed revision first enters the candidate population and is evaluated in subsequent matched rollouts; task rewards update its rating and determine whether it is retained in the final workflow state.

\paragraph{Sparse feedback and adaptation stability.}
A limited target-interaction budget can produce noisy or uniform rewards, making it difficult to distinguish useful context revisions and assign policy credit. To reduce this ambiguity, we evaluate context variants on matched tasks and reset seeds, and compute policy advantages only among rollouts sharing the same task and context condition. Groups without reward variation are excluded from policy updates rather than assigned artificial credit. We also clear the on-policy buffer after each update attempt and restrict training to a LoRA adapter on a frozen backbone, reducing stale-policy updates and limiting parameter drift under sparse supervision.

\paragraph{Deployment safety and continual use.}
Exploration on a new application may expose private information, trigger irreversible actions, or accumulate unstable updates over time. We therefore
conduct adaptation in resettable emulators, restrict interaction to designated adaptation tasks and a fixed budget, and freeze the selected context and LoRA checkpoint before held-out use. App-bound workflow state remains local to its application, while only validated transferable knowledge may cross application boundaries. Updating only a lightweight LoRA adapter on a frozen backbone
further limits parameter drift and the cost of repeated adaptation.

%% file: tex/conclusion.tex
\section{Conclusion}

We introduced \tool, a test-time adaptation framework that updates workflow
context and policy on unseen applications using the agent's own target
rollouts and task rewards. Its transfer-constrained context retains reusable
workflow knowledge while excluding app-bound source state, and its policy
update assigns credit only within matched task--context conditions. \tool
reaches 45.0\% on AndroidWorld-Generalization, compared with 37.5\% for the
reported Policy-Only TTA baseline, and improves AndroidWorld Plus from 38.6\% to
52.9\%. Context adaptation improves performance without policy updates in both settings, while \tool achieves the strongest overall results, supporting context and policy adaptation as distinct but complementary channels for novel-app generalization.

%% file: appendix.tex

\appendix
\onecolumn

\setcounter{secnumdepth}{2}
\setcounter{tocdepth}{2}
\setlength{\parindent}{1em}
\setlength{\parskip}{0pt}
\setlength{\textfloatsep}{10pt plus 2pt minus 2pt}
\setlength{\floatsep}{9pt plus 2pt minus 2pt}
\begin{center}
{\LARGE\bfseries Appendix / Supplementary}
\end{center}
\noindent\textbf{Appendix overview.}
This appendix reports the complete data allocation, implementation details,
workflow-context examples, additional analyses, prompt templates, and
validation contracts for CoAdapt-GUI. Section numbering is continuous within
the appendix and cross-references are resolved in the main document.

\section{AndroidWorld Plus Construction and Data Allocation}
\label{app:awplus-data}

\subsection{Benchmark construction and executable filtering}

AndroidWorld Plus is an experimental extension of
AndroidWorld~\citep{rawles2024androidworld}. We retain the 19 AndroidWorld apps
with app-attributable executable tasks and add six apps with their task suites:
Calculator, Snapseed, and Wikipedia from B-MoCA~\citep{lee2024bmoca}, and
Bluecoins, Maps.me, and Pi Music from AndroidLab~\citep{xu2024androidlab}.
This produces 25 apps spanning 12 Google Play Store categories. We use the
published Play Store category as an external functional taxonomy, following
Android Control~\citep{li2024androidcontrol}, rather than defining categories
from task outcomes.

The merged metadata contains 194 task rows. Two generic or composite tasks,
\texttt{OpenApp} and \texttt{SaveCopyOfReceipt}, cannot be attributed to one
app and are excluded from the app-level split, leaving 192 app-attributable
templates. We then validate every template against the executable task
registry. \texttt{WikipediaDecreaseTextSize50} is defined in the upstream task
file but is not registered with the execution harness and therefore cannot be
instantiated. We remove this template before freezing the manifests, yielding
191 executable task templates. No filtering decision uses an agent rollout,
reward, or evaluation outcome.

\subsection{Source and target app pools}

The source and target app pools are disjoint. The source pool contains 12 apps
and 96 templates (Table~\ref{tab:awplus-source-allocation}). Five task seeds per
template, \(\{30,31,32,33,34\}\), provide 480 source episodes for source-side
policy training and workflow construction.

\begin{table}[!ht]
\footnotesize
\centering
\setlength{\tabcolsep}{5pt}
\begin{tabular*}{\textwidth}{@{\extracolsep{\fill}}llrllr@{}}
\hline
\textbf{Source app} & \textbf{Category} & \textbf{Templates} &
\textbf{Source app} & \textbf{Category} & \textbf{Templates} \\
\hline
\texttt{markor}               & Productivity    & 14 &
\texttt{tasks\_org}           & Productivity    &  6 \\
\texttt{joplin}               & Productivity    &  4 &
\texttt{calculator}           & Tools           & 19 \\
\texttt{clock}                & Tools           &  3 &
\texttt{files}                & Tools           &  2 \\
\texttt{bluecoins}            & Finance         & 15 &
\texttt{pi\_music}            & Music \& Audio  & 12 \\
\texttt{audio\_recorder}      & Music \& Audio  &  2 &
\texttt{snapseed}             & Photography     & 11 \\
\texttt{simple\_sms\_messenger} & Communication &  6 &
\texttt{contacts}             & Communication   &  2 \\
\hline
\multicolumn{5}{r}{\textbf{Total}} & \textbf{96} \\
\hline
\end{tabular*}
\caption{Complete AndroidWorld Plus source-app allocation. All source
templates are excluded from the target pool.}
\label{tab:awplus-source-allocation}
\end{table}

The target pool contains the remaining 13 apps and 95 templates. For the
category-coverage analysis, a target app's group depends only on whether its
category is represented in the source pool. Six target apps satisfy this
condition and form the \textbf{Category-Shared Apps} group. The other seven
belong to categories absent from the source pool and form the
\textbf{Category-Novel Apps} group. Table~\ref{tab:awplus-target-allocation}
gives the complete target allocation.

\begin{table}[!ht]
\footnotesize
\centering
\setlength{\tabcolsep}{4.5pt}
\begin{tabular*}{\textwidth}{@{\extracolsep{\fill}}lllrrrr@{}}
\hline
\textbf{Group} & \textbf{Target app} & \textbf{Category} &
\multicolumn{2}{c}{\textbf{Templates}} &
\multicolumn{2}{c}{\textbf{Episodes}} \\
\cline{4-7}
& & & \textbf{Adapt.} & \textbf{Eval.} &
\textbf{Adapt. pool} & \textbf{Eval.} \\
\hline
Category-Shared & \texttt{simple\_calendar\_pro} & Productivity
& 11 & 6 & 55 & 18 \\
Category-Shared & \texttt{system\_settings} & Tools
& 9 & 6 & 45 & 18 \\
Category-Shared & \texttt{pro\_expense} & Finance
& 6 & 3 & 30 & 9 \\
Category-Shared & \texttt{retro\_music} & Music \& Audio
& 3 & 1 & 15 & 3 \\
Category-Shared & \texttt{camera} & Photography
& 1 & 1 & 5 & 3 \\
Category-Shared & \texttt{chrome} & Communication
& 2 & 1 & 10 & 3 \\
\cline{2-7}
& \multicolumn{2}{l}{\textit{Category-Shared subtotal}}
& 32 & 18 & 160 & 54 \\
\hline
Category-Novel & \texttt{maps\_me} & Maps \& Navigation
& 9 & 6 & 45 & 18 \\
Category-Novel & \texttt{osmand} & Maps \& Navigation
& 2 & 1 & 10 & 3 \\
Category-Novel & \texttt{broccoli} & Food \& Drink
& 8 & 5 & 40 & 15 \\
Category-Novel & \texttt{opentracks} & Health \& Fitness
& 4 & 2 & 20 & 6 \\
Category-Novel & \texttt{wikipedia} & Books \& Reference
& 3 & 2 & 15 & 6 \\
Category-Novel & \texttt{vlc} & Video Players \& Editors
& 1 & 1 & 5 & 3 \\
Category-Novel & \texttt{simple\_draw\_pro} & Art \& Design
& 1 & 0 & 5 & 0 \\
\cline{2-7}
& \multicolumn{2}{l}{\textit{Category-Novel subtotal}}
& 28 & 17 & 140 & 51 \\
\hline
& \multicolumn{2}{l}{\textbf{Target total}}
& \textbf{60} & \textbf{35} & \textbf{300} & \textbf{105} \\
\hline
\end{tabular*}
\caption{Complete AndroidWorld Plus target-app allocation. The adaptation
pool uses five seeds per adaptation template; frozen evaluation uses three
disjoint seeds per evaluation template. \texttt{simple\_draw\_pro} has one
executable template and is used only for adaptation, so the held-out panel
covers 12 of the 13 target apps.}
\label{tab:awplus-target-allocation}
\end{table}

\subsection{Template and seed separation}

Within each target app, we sort the executable template identifiers and create
disjoint adaptation and evaluation sets. For an app with at least three
templates, the first \(\lceil0.6n\rceil\) templates are assigned to adaptation
and the remainder to evaluation. Two-template apps use a one--one split; a
single-template app contributes only to adaptation. This deterministic rule
produces 60 adaptation templates and 35 evaluation templates, with no template
overlap in any target app. The split was frozen before target rollout
collection and does not use task rewards, policy predictions, or evaluation
outcomes.

Each adaptation template is instantiated with seeds
\(\{30,31,32,33,34\}\), defining a 300-instance sampling pool. Each evaluation
template is instantiated with the disjoint seeds \(\{40,41,42\}\), producing
the frozen 105-episode panel used for all reported AndroidWorld Plus results.
The 300 adaptation instances specify the available sampling pool rather than
the number of target rollout calls: CoAdapt-GUI samples from this pool under
the fixed interaction schedule described in
Appendix~\ref{app:implementation}.

\subsection{Complete source and target template manifests}
\label{app:awplus-template-manifest}

Tables~\ref{tab:awplus-source-template-manifest} and
\ref{tab:awplus-target-template-manifest} list all 191 executable
templates in the frozen AndroidWorld Plus manifest. Source templates are
used only for source-side policy training and workflow construction. Target
templates are partitioned into adaptation and held-out evaluation sets
within each target app. Identifiers match the execution harness exactly.

\begingroup
\footnotesize
\setlength{\tabcolsep}{3pt}
\renewcommand{\arraystretch}{1.05}
\setlength{\LTleft}{0pt}
\setlength{\LTright}{0pt}

\begin{longtable}{@{}>{\raggedright\arraybackslash}p{0.14\linewidth}>{\raggedright\arraybackslash}p{0.15\linewidth}>{\centering\arraybackslash}p{0.06\linewidth}>{\raggedright\arraybackslash}p{0.59\linewidth}@{}}
\caption{Complete AndroidWorld Plus source-pool template manifest}\label{tab:awplus-source-template-manifest}\\
\toprule
\textbf{Source app} & \textbf{Category} & \textbf{Count} & \textbf{Executable source templates} \\
\midrule
\endfirsthead
\multicolumn{4}{l}{\textit{Complete AndroidWorld Plus source-pool template manifest (continued)}}\\
\toprule
\textbf{Source app} & \textbf{Category} & \textbf{Count} & \textbf{Executable source templates} \\
\midrule
\endhead
\midrule
\multicolumn{4}{r}{\textit{Continued on the next page}}\\
\endfoot
\bottomrule
\endlastfoot
\texttt{markor} & Productivity &
14 &
\manifesttask{MarkorAddNoteHeader}
\manifesttask{MarkorChangeNoteContent}
\manifesttask{MarkorCreateFolder}
\manifesttask{MarkorCreateNote}
\manifesttask{MarkorCreateNoteAndSms}
\manifesttask{MarkorCreateNoteFromClipboard}
\manifesttask{MarkorDeleteAllNotes}
\manifesttask{MarkorDeleteNewestNote}
\manifesttask{MarkorDeleteNote}
\manifesttask{MarkorEditNote}
\manifesttask{MarkorMergeNotes}
\manifesttask{MarkorMoveNote}
\manifesttask{MarkorTranscribeReceipt}
\manifesttask{MarkorTranscribeVideo} \\
\midrule
\texttt{tasks\_org} & Productivity &
6 &
\manifesttask{TasksCompletedTasksForDate}
\manifesttask{TasksDueNextWeek}
\manifesttask{TasksDueOnDate}
\manifesttask{TasksHighPriorityTasks}
\manifesttask{TasksHighPriorityTasksDueOnDate}
\manifesttask{TasksIncompleteTasksOnDate} \\
\midrule
\texttt{joplin} & Productivity &
4 &
\manifesttask{NotesIsTodo}
\manifesttask{NotesMeetingAttendeeCount}
\manifesttask{NotesRecipeIngredientCount}
\manifesttask{NotesTodoItemCount} \\
\midrule
\texttt{calculator} & Tools &
19 &
\manifesttask{CalculatorConvert45DegreesToRadians}
\manifesttask{CalculatorGeometricMean}
\manifesttask{CalculatorHarmonicMean}
\manifesttask{CalculatorInput1}
\manifesttask{CalculatorInput10Choose2}
\manifesttask{CalculatorInput17Times23}
\manifesttask{CalculatorInput1Plus1}
\manifesttask{CalculatorInput2Plus24Div3}
\manifesttask{CalculatorInput3Times5}
\manifesttask{CalculatorInput5Choose2}
\manifesttask{CalculatorInputCos180}
\manifesttask{CalculatorInputCos60}
\manifesttask{CalculatorInputFactorial6}
\manifesttask{CalculatorInputLn1234}
\manifesttask{CalculatorInputPercent50Of28}
\manifesttask{CalculatorInputSqrt25}
\manifesttask{CalculatorOpen}
\manifesttask{CalculatorSumFirst5Fibonacci}
\manifesttask{CalculatorSumFirst5Primes} \\
\midrule
\texttt{clock} & Tools &
3 &
\manifesttask{ClockStopWatchPausedVerify}
\manifesttask{ClockStopWatchRunning}
\manifesttask{ClockTimerEntry} \\
\midrule
\texttt{files} & Tools &
2 &
\manifesttask{FilesDeleteFile}
\manifesttask{FilesMoveFile} \\
\midrule
\texttt{bluecoins} & Finance &
15 &
\manifesttask{BluecoinsAddExpense}
\manifesttask{BluecoinsAddExpenseOnDate}
\manifesttask{BluecoinsAddExpenseOnDateWithLabel}
\manifesttask{BluecoinsAddIncomeOnDateWithNote}
\manifesttask{BluecoinsAddIncomeWithLabel}
\manifesttask{BluecoinsEditExpenseAmount}
\manifesttask{BluecoinsEditExpenseDateAmountNote}
\manifesttask{BluecoinsEditIncomeDateAndAmount}
\manifesttask{BluecoinsEditTransactionType}
\manifesttask{BluecoinsEditTransactionTypeAmountNote}
\manifesttask{BluecoinsQueryCategorySpending}
\manifesttask{BluecoinsQuerySpendingCategory}
\manifesttask{BluecoinsQuerySpendingOnDate}
\manifesttask{BluecoinsQueryTotalSpendingOnDate}
\manifesttask{BluecoinsQueryTransactionCount} \\
\midrule
\texttt{pi\_music} & Music \& Audio &
12 &
\manifesttask{PiMusicCreatePlaylist}
\manifesttask{PiMusicPauseAndSeek}
\manifesttask{PiMusicPlayFromPlaylist}
\manifesttask{PiMusicPlaySongByTitleArtist}
\manifesttask{PiMusicQueryArtistSongCount}
\manifesttask{PiMusicQueryArtistTotalDuration}
\manifesttask{PiMusicQueryLongestSongDuration}
\manifesttask{PiMusicQuerySongAlbum}
\manifesttask{PiMusicQuerySortedSongsByTitle}
\manifesttask{PiMusicQueryTotalSongs}
\manifesttask{PiMusicSortByDurationAscending}
\manifesttask{PiMusicSortByDurationDescending} \\
\midrule
\texttt{audio\_recorder} & Music \& Audio &
2 &
\manifesttask{AudioRecorderRecordAudio}
\manifesttask{AudioRecorderRecordAudioWithFileName} \\
\midrule
\texttt{snapseed} & Photography &
11 &
\manifesttask{SnapseedTask1}
\manifesttask{SnapseedTask10}
\manifesttask{SnapseedTask11}
\manifesttask{SnapseedTask2}
\manifesttask{SnapseedTask3}
\manifesttask{SnapseedTask4}
\manifesttask{SnapseedTask5}
\manifesttask{SnapseedTask6}
\manifesttask{SnapseedTask7}
\manifesttask{SnapseedTask8}
\manifesttask{SnapseedTask9} \\
\midrule
\texttt{\seqsplit{simple\_sms\_messenger}} & Communication &
6 &
\manifesttask{SimpleSmsReply}
\manifesttask{SimpleSmsReplyMostRecent}
\manifesttask{SimpleSmsResend}
\manifesttask{SimpleSmsSend}
\manifesttask{SimpleSmsSendClipboardContent}
\manifesttask{SimpleSmsSendReceivedAddress} \\
\midrule
\texttt{contacts} & Communication &
2 &
\manifesttask{ContactsAddContact}
\manifesttask{ContactsNewContactDraft} \\
\midrule
\end{longtable}

\scriptsize
\setlength{\tabcolsep}{4pt}

\begin{longtable}{@{}>{\raggedright\arraybackslash}p{0.48\linewidth}>{\raggedright\arraybackslash}p{0.48\linewidth}@{}}
\caption{Complete AndroidWorld Plus target adaptation/evaluation manifest}\label{tab:awplus-target-template-manifest}\\
\toprule
\textbf{Adaptation templates} & \textbf{Held-out evaluation templates} \\
\midrule
\endfirsthead
\multicolumn{2}{l}{\textit{Complete AndroidWorld Plus target adaptation/evaluation manifest (continued)}}\\
\toprule
\textbf{Adaptation templates} & \textbf{Held-out evaluation templates} \\
\midrule
\endhead
\midrule
\multicolumn{2}{r}{\textit{Continued on the next page}}\\
\endfoot
\bottomrule
\endlastfoot
\multicolumn{2}{@{}l@{}}{\textbf{\texttt{simple\_calendar\_pro}}\quad \textit{Category-Shared; Productivity}} \\*
\manifestcount{11}\manifesttask{SimpleCalendarAddOneEvent}
\manifesttask{SimpleCalendarAddOneEventInTwoWeeks}
\manifesttask{SimpleCalendarAddOneEventRelativeDay}
\manifesttask{SimpleCalendarAddOneEventTomorrow}
\manifesttask{SimpleCalendarAddRepeatingEvent}
\manifesttask{SimpleCalendarAnyEventsOnDate}
\manifesttask{SimpleCalendarDeleteEvents}
\manifesttask{SimpleCalendarDeleteEventsOnRelativeDay}
\manifesttask{SimpleCalendarDeleteOneEvent}
\manifesttask{SimpleCalendarEventOnDateAtTime}
\manifesttask{SimpleCalendarEventsInNextWeek} &
\manifestcount{6}\manifesttask{SimpleCalendarEventsInTimeRange}
\manifesttask{SimpleCalendarEventsOnDate}
\manifesttask{SimpleCalendarFirstEventAfterStartTime}
\manifesttask{SimpleCalendarLocationOfEvent}
\manifesttask{SimpleCalendarNextEvent}
\manifesttask{SimpleCalendarNextMeetingWithPerson} \\
\midrule
\multicolumn{2}{@{}l@{}}{\textbf{\texttt{system\_settings}}\quad \textit{Category-Shared; Tools}} \\*
\manifestcount{9}\manifesttask{SystemBluetoothTurnOff}
\manifesttask{SystemBluetoothTurnOffVerify}
\manifesttask{SystemBluetoothTurnOn}
\manifesttask{SystemBluetoothTurnOnVerify}
\manifesttask{SystemBrightnessMax}
\manifesttask{SystemBrightnessMaxVerify}
\manifesttask{SystemBrightnessMin}
\manifesttask{SystemBrightnessMinVerify}
\manifesttask{SystemCopyToClipboard} &
\manifestcount{6}\manifesttask{SystemWifiTurnOff}
\manifesttask{SystemWifiTurnOffVerify}
\manifesttask{SystemWifiTurnOn}
\manifesttask{SystemWifiTurnOnVerify}
\manifesttask{TurnOffWifiAndTurnOnBluetooth}
\manifesttask{TurnOnWifiAndOpenApp} \\
\midrule
\multicolumn{2}{@{}l@{}}{\textbf{\texttt{pro\_expense}}\quad \textit{Category-Shared; Finance}} \\*
\manifestcount{6}\manifesttask{ExpenseAddMultiple}
\manifesttask{ExpenseAddMultipleFromGallery}
\manifesttask{ExpenseAddMultipleFromMarkor}
\manifesttask{ExpenseAddSingle}
\manifesttask{ExpenseDeleteDuplicates}
\manifesttask{ExpenseDeleteDuplicates2} &
\manifestcount{3}\manifesttask{ExpenseDeleteMultiple}
\manifesttask{ExpenseDeleteMultiple2}
\manifesttask{ExpenseDeleteSingle} \\
\midrule
\multicolumn{2}{@{}l@{}}{\textbf{\texttt{retro\_music}}\quad \textit{Category-Shared; Music \& Audio}} \\*
\manifestcount{3}\manifesttask{RetroCreatePlaylist}
\manifesttask{RetroPlayingQueue}
\manifesttask{RetroPlaylistDuration} &
\manifestcount{1}\manifesttask{RetroSavePlaylist} \\
\midrule
\multicolumn{2}{@{}l@{}}{\textbf{\texttt{camera}}\quad \textit{Category-Shared; Photography}} \\*
\manifestcount{1}\manifesttask{CameraTakePhoto} &
\manifestcount{1}\manifesttask{CameraTakeVideo} \\
\midrule
\multicolumn{2}{@{}l@{}}{\textbf{\texttt{chrome}}\quad \textit{Category-Shared; Communication}} \\*
\manifestcount{2}\manifesttask{BrowserDraw}
\manifesttask{BrowserMaze} &
\manifestcount{1}\manifesttask{BrowserMultiply} \\
\midrule
\multicolumn{2}{@{}l@{}}{\textbf{\texttt{maps\_me}}\quad \textit{Category-Novel; Maps \& Navigation}} \\*
\manifestcount{9}\manifesttask{MapsMeAddWorkPlace}
\manifesttask{MapsMeCheckDrivingDistanceTime}
\manifesttask{MapsMeCheckNearestHotel}
\manifesttask{MapsMeCheckNearestPlace}
\manifesttask{MapsMeCheckNearestPlaceDriveTime}
\manifesttask{MapsMeCheckNearestPlaceWalkTime}
\manifesttask{MapsMeCheckPublicTransportRoute}
\manifesttask{MapsMeCheckRidingTime}
\manifesttask{MapsMeCheckWalkingDistanceTime} &
\manifestcount{6}\manifesttask{MapsMeCompareRidingVsPublicTransport}
\manifesttask{MapsMeNavigateToBerkeley}
\manifesttask{MapsMeNavigateToLocation}
\manifesttask{MapsMeNavigateToOpenAI}
\manifesttask{MapsMeNavigateToStanford}
\manifesttask{MapsMeNavigateToUniversitySouth} \\
\midrule
\multicolumn{2}{@{}l@{}}{\textbf{\texttt{osmand}}\quad \textit{Category-Novel; Maps \& Navigation}} \\*
\manifestcount{2}\manifesttask{OsmAndFavorite}
\manifesttask{OsmAndMarker} &
\manifestcount{1}\manifesttask{OsmAndTrack} \\
\midrule
\multicolumn{2}{@{}l@{}}{\textbf{\texttt{broccoli}}\quad \textit{Category-Novel; Food \& Drink}} \\*
\manifestcount{8}\manifesttask{RecipeAddMultipleRecipes}
\manifesttask{RecipeAddMultipleRecipesFromImage}
\manifesttask{RecipeAddMultipleRecipesFromMarkor}
\manifesttask{RecipeAddMultipleRecipesFromMarkor2}
\manifesttask{RecipeAddSingleRecipe}
\manifesttask{RecipeDeleteDuplicateRecipes}
\manifesttask{RecipeDeleteDuplicateRecipes2}
\manifesttask{RecipeDeleteDuplicateRecipes3} &
\manifestcount{5}\manifesttask{RecipeDeleteMultipleRecipes}
\manifesttask{RecipeDeleteMultipleRecipesWithConstraint}
\manifesttask{RecipeDeleteMultipleRecipesWithNoise}
\manifesttask{RecipeDeleteSingleRecipe}
\manifesttask{RecipeDeleteSingleWithRecipeWithNoise} \\
\midrule
\multicolumn{2}{@{}l@{}}{\textbf{\texttt{opentracks}}\quad \textit{Category-Novel; Health \& Fitness}} \\*
\manifestcount{4}\manifesttask{SportsTrackerActivitiesCountForWeek}
\manifesttask{SportsTrackerActivitiesOnDate}
\manifesttask{SportsTrackerActivityDuration}
\manifesttask{SportsTrackerLongestDistanceActivity} &
\manifestcount{2}\manifesttask{SportsTrackerTotalDistanceForCategoryOverInterval}
\manifesttask{SportsTrackerTotalDurationForCategoryThisWeek} \\
\midrule
\multicolumn{2}{@{}l@{}}{\textbf{\texttt{wikipedia}}\quad \textit{Category-Novel; Books \& Reference}} \\*
\manifestcount{3}\manifesttask{WikipediaDisablePreviewAndFeed}
\manifesttask{WikipediaGoToSavedTab}
\manifesttask{WikipediaGoToSearchTab} &
\manifestcount{2}\manifesttask{WikipediaIncreaseTextSize180}
\manifesttask{WikipediaOpen} \\
\midrule
\multicolumn{2}{@{}l@{}}{\textbf{\texttt{vlc}}\quad \textit{Category-Novel; Video Players \& Editors}} \\*
\manifestcount{1}\manifesttask{VlcCreatePlaylist} &
\manifestcount{1}\manifesttask{VlcCreateTwoPlaylists} \\
\midrule
\multicolumn{2}{@{}l@{}}{\textbf{\texttt{simple\_draw\_pro}}\quad \textit{Category-Novel; Art \& Design}} \\*
\manifestcount{1}\manifesttask{SimpleDrawProCreateDrawing} &
\manifestcount{0}\emph{None} \\
\midrule
\end{longtable}
\endgroup
\section{AndroidWorld-Generalization Protocol and Run Accounting}
\label{app:awgen-protocol}

\subsection{Released source and target splits}

Setting~1 follows the released AndroidWorld-Generalization unseen-app
split~\citep{gu2026generalization}. Its source split contains 12 apps, 62 task
templates, and 905 training instances. Every configuration starts from the
released step-500 UI-TARS-7B checkpoint~\citep{qin2025uitarspioneeringautomatedgui},
which was trained only on this source split. The reusable source workflow
library used by CoAdapt-GUI is likewise constructed only from this source
split.

The target split contains five apps absent from the source split. It provides
eight adaptation instances per app, for 40 in total, and 48 held-out evaluation
instances. Adaptation and evaluation cover the same 16 task templates but
instantiate them with non-overlapping seeds. This setting therefore evaluates
transfer to new instances of task types encountered during target adaptation,
rather than transfer to new task templates.

\begin{table}[!ht]
\footnotesize
\centering
\setlength{\tabcolsep}{4pt}
\begin{tabular}{@{}p{0.13\textwidth}p{0.55\textwidth}rr@{}}
\hline
\textbf{Target app} & \textbf{Task templates (released task ID)} &
\textbf{Adapt. inst.} & \textbf{Eval. inst.} \\
\hline
Audio Recorder &
\texttt{AudioRecorderRecordAudioWithFileName} (28) & 8 & 3 \\
Clock &
\texttt{ClockTimerEntry} (20) & 8 & 3 \\
OsmAnd &
\texttt{OsmAndFavorite} (74); \texttt{OsmAndMarker} (88) & 8 & 6 \\
Tasks &
\texttt{TasksDueOnDate} (100); \texttt{TasksHighPriorityTasksDueOnDate}
(102); \texttt{TasksCompletedTasksForDate} (104);
\texttt{TasksIncompleteTasksOnDate} (105) & 8 & 12 \\
Broccoli &
\texttt{\seqsplit{RecipeDeleteMultipleRecipes}} (3);
\texttt{RecipeDeleteSingleRecipe} (4);
\texttt{RecipeAddSingleRecipe} (31);
\texttt{\seqsplit{RecipeDeleteSingleWithRecipeWithNoise}} (32);
\texttt{RecipeAddMultipleRecipes} (56);
\texttt{\seqsplit{RecipeDeleteMultipleRecipesWithNoise}} (57);
\texttt{RecipeAddMultipleRecipesFromMarkor2} (82); and
\texttt{\seqsplit{RecipeDeleteMultipleRecipesWithConstraint}} (90) & 8 & 24 \\
\hline
\textbf{Total} & \textbf{16 templates} & \textbf{40} & \textbf{48} \\
\hline
\end{tabular}
\caption{Complete target allocation for AndroidWorld-Generalization. Each
template is present in both splits, but the concrete task instances are
disjoint.}
\label{tab:awgen-target-allocation}
\end{table}

\subsection{Instance seeds and leakage control}

The released evaluation manifest instantiates every target template with seeds
\(\{7,30,1234\}\). Adaptation uses different seeds: Audio Recorder and Clock
use \(\{1,2,3,4,5,6,8,9\}\); each OsmAnd template uses
\(\{1,2,3,4\}\); each Tasks template contributes two instances; and each
Broccoli template contributes one instance. The released Tasks instances use
seeds \(\{1,2\}\), except
\texttt{TasksCompletedTasksForDate}, which uses \(\{2,3\}\). Seven Broccoli
templates use seed 1 and
\texttt{RecipeDeleteSingleWithRecipeWithNoise} uses seed 2. Consequently, the
intersection between adaptation and evaluation
\((\texttt{task\_id},\texttt{seed})\) pairs is empty.

All context revisions, policy updates, and checkpoint selection use only the
40-instance adaptation manifest. The workflow context and policy adapter are
then frozen before the 48-instance evaluation manifest is opened. No held-out
task instance, trajectory, reward, or outcome is used for adaptation or model
selection.

\subsection{Update schedule and result provenance}

The adaptive configurations use the released eight adaptation instances per
target app and a nominal 50-update schedule per app. They are evaluated on the
same frozen 48-instance manifest. The schedule fixes the number of method-level
updates; because context revision and policy optimization use different update
operators, it should not be interpreted as an identical gradient or compute
budget across methods.

The Base Policy and Policy-Only TTA values in the main paper are reported
results from AndroidWorld-Generalization~\citep{gu2026generalization} and were
not reproduced in our environment. Static Context Transfer, Context-Only TTA,
and CoAdapt-GUI are our results. Each is evaluated in five complete runs, and
the main paper reports the mean and standard deviation over those runs. All
five runs use the same frozen adaptation and evaluation manifests; no
incomplete run is included in the reported statistics.

\section{CoAdapt-GUI Implementation Details}
\label{app:implementation}

This section specifies the realization of the context and policy branches in
Algorithm~1 of the main paper. It describes CoAdapt-GUI itself rather than
restating the implementation of every comparison configuration. The notation
\texttt{layer1} and \texttt{layer2} below refers to the runtime JSON schema:
\texttt{layer1} is the app-bound screen-transition state, and
\texttt{layer2} is the transferable workflow state described in the paper.

\subsection{Policy and source initialization}

For AndroidWorld-Generalization, CoAdapt-GUI starts from the released step-500
UI-TARS-7B checkpoint and uses the released 50-step target schedule described
in Appendix~\ref{app:awgen-protocol}. For AndroidWorld Plus, the acting policy
is Qwen3-VL-8B-Instruct at revision
\texttt{0c351dd01ed87e9c1b53cbc748cba10e6187ff3b}. The VLM backbone is loaded
in bfloat16 and remains frozen. We initialize it with a rank-16 LoRA adapter
trained only on the 12 source apps. All configurations within a matched
comparison use the same source-trained initialization, and configurations
without policy adaptation keep it frozen. No target rollout, reward, or
held-out outcome is used to construct or select this initialization.

Source workflow construction is independent of source-side LoRA training. A
frozen synthesizer converts audited source trajectories into one app-bound
screen-transition state and one transferable workflow state per source app.
Only validated transferable entries are grouped by the externally defined Play
Store category and consolidated into the immutable source library. App-bound
screen states, transitions, UI strings, and identifiers never enter this
library.

Table~\ref{tab:source-library-composition} reports the resulting frozen
library. A target app can retrieve only the single library matching its Play
Store category; cross-category retrieval is disabled. Thus a Category-Novel
App receives no source entry, rather than an entry selected by semantic
similarity or target reward.

\begin{table}[!ht]
\small
\centering
\setlength{\tabcolsep}{8pt}
\begin{tabular}{@{}lrr@{}}
\hline
\textbf{Source category} & \textbf{Source apps} & \textbf{Workflow entries} \\
\hline
Productivity & 3 & 14 \\
Tools & 3 & 15 \\
Finance & 1 & 7 \\
Music \& Audio & 2 & 11 \\
Photography & 1 & 6 \\
Communication & 2 & 7 \\
\hline
\textbf{Total} & \textbf{12} & \textbf{60} \\
\hline
\end{tabular}
\caption{Composition of the frozen transferable source-workflow library used
by AndroidWorld Plus. Counts refer to validated structured workflow entries,
not source trajectories or task templates.}
\label{tab:source-library-composition}
\end{table}

\subsection{AndroidWorld Plus adaptation configuration}

Table~\ref{tab:implementation-hyperparameters} records the final configuration
used for the AndroidWorld Plus experiments. The four rollouts associated with
one task--context condition form the group size $N=4$; $N$ is not the
number of context variants. At most $M=2$ already materialized variants are
selected in a round, so 20 rounds require at most
$20\times2\times4=160$ target rollout calls per app. Early rounds may use
only the root variant and therefore consume fewer calls.

\begin{table}[t]
\footnotesize
\centering
\setlength{\tabcolsep}{4pt}
\begin{tabular}{@{}p{0.18\textwidth}p{0.43\textwidth}p{0.31\textwidth}@{}}
\hline
\textbf{Component} & \textbf{Parameter} & \textbf{Final value} \\
\hline
Policy & Backbone / numerical precision & Qwen3-VL-8B-Instruct / bfloat16 \\
Policy & Maximum image pixels / generated tokens & $1{,}605{,}632$ / 512 \\
Policy & LoRA rank / scale / configured dropout & $16 / 32 / 0.0$ \\
Policy & LoRA target modules & \texttt{q\_proj}, \texttt{v\_proj} \\
Policy & Optimizer / learning rate & AdamW / $3\times10^{-4}$ \\
Policy & Maximum gradient norm & $1.0$ \\
Policy & Advantage normalization & mean-centered within task--context group \\
Policy & Minimum active trajectories for an update & $3$ \\
Policy & Frozen-policy anchor coefficient / log-ratio clip & $0.05 / 10.0$ \\
\hline
Context & Initial TrueSkill mean / standard deviation & $25.0 / 8.33$ \\
Context & TrueSkill performance / dynamics parameters & $4.17 / 0.083$ \\
Context & Population window / child uncertainty increment & $15 / 1.5$ \\
Context & Selection optimism / softmax temperature & $1.0 / 1.0$ \\
Context & Frozen synthesizer and reflector & Claude Opus 4.7 \\
Context & Reflection / diff output limits & $4{,}096 / 32{,}000$ tokens \\
Context & Malformed-output retries & $3$ \\
\hline
Protocol & Adaptation rounds per target app & $20$ \\
Protocol & Context variants per round & $M\leq2$ \\
Protocol & Rollouts per task--context condition & $N=4$ \\
Protocol & Maximum rollout calls per app & $160$ \\
Protocol & Adaptation / evaluation task seeds &
\(\{30,31,32,33,34\} / \{40,41,42\}\) \\
Protocol & Hardware / mean adaptation time & NVIDIA H200 141 GB / 9--10 GPU-hours \\
\hline
\end{tabular}
\caption{CoAdapt-GUI implementation details for AndroidWorld Plus. The policy
buffer is transient and is cleared after every update attempt; its lifetime is
specified below.}
\label{tab:implementation-hyperparameters}
\end{table}

\subsection{Context population and reward-guided revision}

Each target app has an independent population of transferable workflow states.
For a Category-Shared App, the root contains the retrieved category-matched
source workflow context. For a Category-Novel App, source retrieval returns
empty and the target-grounded root is initialized empty. In the latter case,
the first validated child constructs an initial transferable entry from target
rollout evidence; later children refine that entry in the same way as in the
nonempty case.

At round $t$, the controller samples up to two variants from the latest
15-member population window. Sampling is proportional to

\[
p(i) \propto
\exp\!\left(\frac{\mu_i+\lambda\sigma_i}{T}\right),
\qquad \lambda=1,\quad T=1,
\]

where \((\mu_i,\sigma_i)\) is the variant's TrueSkill rating. Selected
variants are evaluated on matched task and reset conditions. Their benchmark
task rewards determine their ordering and update their ratings. A frozen
reflector then contrasts the resulting success and failure traces and proposes
a typed revision to a high-rated parent.

The temporal order is important. Only variants that existed before the current
rollouts receive ratings from those rollouts. A newly proposed child is added
only after typed-diff parsing, transferable-state scope filtering, and a nonempty state
change; the caller records its parent, task, iteration, and revision
provenance. The child receives no inherited reward. Its usefulness is measured
only if it is sampled in a later round. At budget exhaustion, the highest-rated
validated state is selected as $M^\star$.

The reflection record contains the task template and instantiated goal, reset
seed, executable task reward, and a compact trace of actions, action reasons,
post-action summaries, UI evidence, and parse or execution status. The caller
retains at most ten UI-element lines per step, clips each line to 160
characters, and clips reasons and summaries to 400 characters. These limits
bound the reflector input without replacing executable rewards with model
judgments.

\subsection{Task--context-matched LoRA update}

Every trajectory is stored with its task, rendered-context identity, reset
identity, and the exact multimodal inputs and action tokens used during
collection. Policy advantages are computed only within groups whose members
share the same task and context variant. In the reported AndroidWorld Plus
configuration, each such group contains four rollouts and uses the
mean-centered advantage

\[
A_j=r_j-\frac{1}{|G|}\sum_{i\in G}r_i.
\]

Singleton or constant-reward groups have zero advantage and are excluded from
the active policy batch. An update is attempted only when at least three active
trajectories remain. For each active trajectory, action-token log-probabilities
are averaged within the trajectory before weighting by $A_j$, preventing a
long rollout from dominating solely because it contains more action units.

The policy buffer is a transient on-policy accumulation buffer, not a replay
memory. The policy remains fixed while a matched batch is collected. After the
batch becomes eligible, at most one LoRA update is attempted and the buffer is
cleared whether or not the batch contains an active reward comparison. No
trajectory is reused after the policy changes. The frozen source-trained
adapter supplies the policy anchor; the numerical log-ratio is clipped to
\([-10,10]\), and the final gradient norm is clipped to 1.0.
Adapter dropout is disabled during both collection and gradient replay so that
the saved behavior probabilities and the replayed current-policy probabilities
refer to the same stochastic policy. Policy-sampling seeds, rather than
dropout, provide the within-condition rollout diversity.

\subsection{Frozen evaluation}

After the interaction budget is exhausted, the selected workflow state and
final LoRA adapter are written as immutable artifacts. Evaluation loads only
these artifacts and the held-out manifest; the reflector, population
controller, and optimizer are disabled. The structural, transferability, and
behavioral checks applied before this point are specified together with the
operative prompts in Appendix~\ref{app:prompts}.

\section{Workflow-Context Representation}
\label{app:workflow-representation}

This section gives concrete examples of the structured context described in
the main paper. Each transferable workflow entry follows
$w=\langle c,P,F,V\rangle$: an applicability condition, an abstract procedure,
failure or recovery conditions, and completion checks. Each example below is
condensed from a validated workflow artifact by removing repeated safeguards
and provenance metadata; no new procedural, failure, or verification rule is
introduced. App names are shown in the headings to identify the originating
run, but are not part of the transferable text injected into the acting policy.

\subsection{App-bound and transferable components}

Table~\ref{tab:representation-boundary} illustrates the transfer boundary with
a source artifact from \texttt{markor}. The app-bound component is a genuine
screen-transition FSM: it includes concrete state names, visible strings,
resource hints, and action-conditioned transitions. The transferable component
retains the task logic while removing those interface bindings. Under unseen-app
adaptation, only the right-hand form is eligible to initialize another app.

\begin{table}[!ht]
\small
\centering
\setlength{\tabcolsep}{6pt}
\begin{tabularx}{\textwidth}{@{}>{\raggedright\arraybackslash}X>{\raggedright\arraybackslash}X@{}}
\hline
\textbf{App-bound screen-transition state (not transferred)} &
\textbf{Transferable workflow entry (eligible)} \\
\hline
\textbf{States:} \texttt{file\_browser}, identified by the Markor title, a red
bottom-right create button, and the Files/To-Do/QuickNote/More navigation;
\texttt{new\_file\_dialog}, identified by the pre-filled
\texttt{my\_note} field, format selector, and FOLDER/CANCEL/OK controls; and
\texttt{editor\_edit\_mode}, identified by the editor toolbar and save icon.

\textbf{Transitions:}\par\smallskip
{\scriptsize\setlength{\tabcolsep}{2pt}
\begin{tabularx}{\linewidth}{@{}>{\raggedright\arraybackslash}Xl@{}}
\textnormal{\textbf{State transition}} & \textnormal{\textbf{Action}} \\
\texttt{file\_browser} $\rightarrow$ \texttt{new\_file\_dialog} &
\texttt{click(fab\_plus)} \\
\texttt{new\_file\_dialog} $\rightarrow$ \texttt{editor\_edit\_mode} &
\texttt{click(OK)}
\end{tabularx}}
&
\textbf{Applicability:} a list or index exposes an affordance for creating an
item.

\textbf{Procedure:} invoke the create affordance; clear any placeholder;
provide the required identifier; select the requested item type if needed;
confirm creation; enter the requested content; and persist the change.

\textbf{Failures:} retaining a placeholder in the identifier, choosing the
wrong item type, creating in the wrong parent container, or declaring success
before persistence.

\textbf{Verification:} the new item appears under the expected parent with the
exact identifier, and reopening it shows the requested content. \\
\hline
\end{tabularx}
\caption{Concrete separation between app-bound interaction state and an
app-agnostic workflow entry. The left column may support execution within its
source app but is excluded from cross-application retrieval.}
\label{tab:representation-boundary}
\end{table}

\subsection{Representative highly rated target contexts}

We next show three entries from highly rated validated states produced during
AndroidWorld Plus adaptation. They cover both initialization regimes: the
calendar example refines retrieved source knowledge for a Category-Shared App,
whereas the navigation and activity-history examples were constructed from
target rollouts after Category-Novel retrieval returned an empty source state.
These entries are selected to illustrate the representation, not as additional
quantitative evaluation examples.

\begin{tcolorbox}[
  enhanced,breakable,
  title={Example 1: Scheduled-entry creation (Category-Shared)},
  colback=PromptBody,colframe=PromptBorder,colbacktitle=PromptTitle,
  coltitle=white,fonttitle=\sffamily\bfseries\small,fontupper=\small,
  boxrule=0.65pt,arc=1.8mm,left=4.5mm,right=4.5mm,top=3.5mm,bottom=3.5mm]
\textbf{Origin.} The target app is \texttt{simple\_calendar\_pro}; its
Productivity category is represented in the source pool. The entry starts from
retrieved source workflow knowledge and is revised using target-app traces.

\textbf{Applicability ($c$).} A create form exposes multiple structured fields,
such as title, date, start time, end time or duration, and optional content.

\textbf{Procedure ($P$).} Enumerate every goal-specified field before acting;
resolve relative dates to absolute dates; set and visually confirm each field;
after confirming the start time, derive the dependent end time; verify which
picker component is active before editing it; save; then reopen the created
item and read back every requested field.

\textbf{Failure/recovery ($F$).} Do not assume that an end-time field inherits
an edited start time. Do not mistake an hour-to-minute auto-advance within one
picker for a transition to another field. If the reasoning and action summary
contain contradictory numeric values, re-inspect the form before continuing.

\textbf{Verification ($V$).} The saved detail view, rather than a potentially
truncated list row, must show the requested date, start and end times, and other
specified fields before completion is emitted.
\end{tcolorbox}

\begin{tcolorbox}[
  enhanced,breakable,
  title={Example 2: Nearest-place and route lookup (Category-Novel)},
  colback=PromptBody,colframe=PromptBorder,colbacktitle=PromptTitle,
  coltitle=white,fonttitle=\sffamily\bfseries\small,fontupper=\small,
  boxrule=0.65pt,arc=1.8mm,left=4.5mm,right=4.5mm,top=3.5mm,bottom=3.5mm]
\textbf{Origin.} The target app is \texttt{maps\_me}; Maps \& Navigation is
absent from the source pool. Its root transferable state is empty, and the
entry is synthesized and subsequently refined from target adaptation traces.

\textbf{Applicability ($c$).} The task asks for a nearby place, place details,
or a route with a specified travel mode.

\textbf{Procedure ($P$).} Establish a valid location and populated map view
before searching; clear the query field and use a canonical place-category
term; inspect both result rows and map markers; open a candidate's detail view;
for routing, set both endpoints and the requested travel mode before reading
the rendered distance or duration.

\textbf{Failure/recovery ($F$).} A viewport-scoped empty result is not global
infeasibility. Recenter or widen the map before reformulating the query. Clear
existing text before retrying to avoid concatenated queries, and avoid random
panning without a location reference.

\textbf{Verification ($V$).} A location reference and populated map are visible;
the selected result's type matches the goal; and any reported place, distance,
or duration is read from the rendered result or route summary.
\end{tcolorbox}

\begin{tcolorbox}[
  enhanced,breakable,
  title={Example 3: Historical activity aggregation (Category-Novel)},
  colback=PromptBody,colframe=PromptBorder,colbacktitle=PromptTitle,
  coltitle=white,fonttitle=\sffamily\bfseries\small,fontupper=\small,
  boxrule=0.65pt,arc=1.8mm,left=4.5mm,right=4.5mm,top=3.5mm,bottom=3.5mm]
\textbf{Origin.} The target app is \texttt{opentracks}; Health \& Fitness is
absent from the source pool. The selected workflow descends from a child that
constructed the category entry from an empty root.

\textbf{Applicability ($c$).} A history query specifies a time window, an
activity type, and an aggregation such as count, total, maximum, or average.

\textbf{Procedure ($P$).} Anchor relative dates to an observed current date;
resolve the requested time interval and week convention; use available date and
activity filters; inspect the full in-window set; disambiguate activity types
using structured metadata rather than user-authored titles; aggregate the
confirmed entries; and format the result in the requested unit and precision.

\textbf{Failure/recovery ($F$).} Avoid first-match answers for maximum or total
queries, title-based activity classification, confusion between calendar weeks
and rolling seven-day windows, and unit conversion based on rounded summary
values when a precise detail value is available.

\textbf{Verification ($V$).} Every included entry has an observed type and
timestamp inside the requested interval; the list boundary has been reached;
units and time formats have been checked; and exactly one answer is followed by
one completion action.
\end{tcolorbox}

\subsection{A recorded context-evolution step}

The Chrome adaptation log provides a compact example of reward-guided context
selection. The retrieved Communication context initially contained generic
navigation and repeated-action procedures. A child proposed after target traces
added three pieces of guidance: handle first-launch gates before continuing the
task, verify that a post-action observation has actually advanced, and change
strategy after repeated no-op actions. Table~\ref{tab:chrome-evolution-event}
shows that the child was admitted without inheriting its parent's reward and
was evaluated only in later rounds.

\begin{table}[!ht]
\small
\centering
\setlength{\tabcolsep}{5pt}
\begin{tabularx}{\textwidth}{@{}p{0.10\textwidth}p{0.18\textwidth}X@{}}
\hline
\textbf{Round} & \textbf{Observed reward} & \textbf{Context event} \\
\hline
6 & root: 0.00 & \texttt{gen1\_mut\_1} is proposed, validated, and admitted
without an assigned reward. \\
7--11 & all: 0.00 & The child and existing alternatives are tested, but every
evaluated condition remains at zero. \\
12 & child: 0.25; root: 0.00 & Under the matched \texttt{BrowserMaze}
condition, the evolved context exposes nonzero executable feedback absent under
the root. \\
\hline
\end{tabularx}
\caption{A recorded Chrome context-evolution event. Values are mean executable
task rewards over the four rollouts for the indicated task--context condition.}
\label{tab:chrome-evolution-event}
\end{table}

The round-12 rating update favored the evolved branch, and later revisions
retained its onboarding and state-advancement checks. The trace illustrates the
intended temporal separation: a typed revision is first validated, then tested
in subsequent rollouts, and retained according to executable task feedback.

\section{Additional Analyses and Ablations}
\label{app:additional-analysis}

\subsection{Incremental contributions of the adaptation channels}

Table~\ref{tab:channel-increments} rewrites the AndroidWorld Plus results from
the main paper as increments between the configurations that isolate each
adaptation channel. Static Context Transfer improves Category-Shared Apps by
9.3 percentage points but leaves Category-Novel Apps unchanged, as expected
from the category-restricted retrieval rule. Policy-Only TTA provides a 1.4
point overall gain but decreases Category-Novel performance. In contrast,
target-side context construction improves both groups over Static Context
Transfer, and enabling the complete joint procedure adds a further 4.8 points
overall over Context-Only TTA.

\begin{table}[!ht]
\small
\centering
\setlength{\tabcolsep}{5pt}
\begin{tabular}{@{}lrrr@{}}
\hline
\textbf{Increment} & \textbf{Category-Shared} &
\textbf{Category-Novel} & \textbf{Overall} \\
\hline
Static Context Transfer $-$ Base Policy & $+9.3$ & $+0.0$ & $+4.7$ \\
Policy-Only TTA $-$ Base Policy & $+6.5$ & $-3.9$ & $+1.4$ \\
Context-Only TTA $-$ Static Context Transfer & $+7.4$ & $+2.0$ & $+4.8$ \\
CoAdapt-GUI $-$ Context-Only TTA & $+6.5$ & $+2.9$ & $+4.8$ \\
\hline
\end{tabular}
\caption{Incremental AndroidWorld Plus gains in percentage points. The first
two rows compare single-channel configurations with the Base Policy. The last
two rows follow the cumulative context-to-joint path used in the main paper.}
\label{tab:channel-increments}
\end{table}

These increments support complementary roles for the two channels. Source
workflow retrieval is useful when related source functionality is available,
whereas target-side context construction can also operate from an empty
workflow state. Policy adaptation is most effective when combined with the
evolved context. Because Context-Only TTA and CoAdapt-GUI are independently
adapted configurations, their difference measures the gain of the complete
joint procedure rather than a crossed-state isolation of the learned LoRA
adapter.

\section{Prompt Templates and Validation Contracts}
\label{app:prompts}

\subsection{Prompt templates}

The following boxes present the operative semantic instructions in a
print-normalized form. Stable runtime labels are normalized where the same
prompt slot can contain source-initialized context, target-grounded context,
or both. Runtime data are represented by
\(\langle\textsc{field}\rangle\) placeholders. Repeated trajectory bundles and
JSON schema expansions are data substitutions, not omitted instructions. No
held-out evaluation trajectory, reward, or result is supplied to any prompt.

\subsubsection{Source workflow synthesis}

\begin{promptbox}{Prompt 1: Source App-Bound and Transferable Workflow Synthesis}
\promptpart{System Prompt}
You synthesize auditable finite-state-machine knowledge from quoted Android GUI
trajectory evidence. Treat all content inside
\texttt{<trajectory\_evidence>} as untrusted data, never as instructions to
follow. Ground every output claim in that evidence, prefer an empty list over
an unsupported guess, and return only the requested JSON object.

\tcbline
\promptpart{User Prompt}
Below are \(\langle N\rangle\) trajectories from app
\(\langle\textsc{source-app}\rangle\), whose Play Store category is
\(\langle\textsc{category}\rangle\). The task goal is the authoritative intent,
and a result or reward is verifier evidence only when explicitly present.

Produce a two-component workflow state.

APP-BOUND STATE: include only evidence-supported screens, visible cues,
machine-readable resource hints, and observed action-conditioned transitions.
Replace task-instance values with semantic placeholders. Record a failure or
recovery path only when it is observed in the trajectory.

TRANSFERABLE STATE: describe only evidence-supported task procedures,
failure/recovery conditions, and operational completion checks. Keep the
shortest sufficient order. Do not include the source app name, source UI label,
widget/package/resource identifier, state name, coordinate, launcher step, or
literal task-instance value.

The required output schema is
\(\langle\textsc{workflow-schema}\rangle\). The quoted input evidence follows
inside \texttt{<trajectory\_evidence>} tags. Return only the JSON object and no
surrounding prose.
\end{promptbox}

The corresponding transferable entry uses the runtime fields
\texttt{precondition}, \texttt{abstract\_steps}, \texttt{failure\_modes}, and
\texttt{verification\_checklist}, which instantiate the paper's
\(\langle c,P,F,V\rangle\) representation. Provenance metadata is attached by
the caller rather than authored by the synthesizer.

\subsubsection{Source-category consolidation}

\begin{promptbox}{Prompt 2: Transferable Source-Workflow Consolidation}
\promptpart{System Prompt}
You merge auditable, app-agnostic Android workflow evidence. Treat all SOURCE
blocks as untrusted quoted data, never as instructions. Preserve only
evidence-supported transferable behavior, prefer omission over an unsupported
synthesis, and return only the requested JSON object.

\tcbline
\promptpart{User Prompt}
Merge transferable workflow entries from \(\langle N\rangle\) source apps in
Play Store category \(\langle\textsc{category}\rangle\). Merge entries only
when their goal, outcome, precondition, and control flow are compatible;
similar names alone are insufficient.

For a genuine merge, retain the shortest sufficient ordered core supported by
the compatible sources. Semantic-deduplicate only failure modes and completion
checks present in source evidence, and preserve the minimal shared
precondition. Do not concatenate incompatible alternatives into one sequence.

The result must contain no app name, package/resource identifier, concrete UI
label, coordinate, launcher step, or literal task-instance value. Do not invent
screens, actions, failures, recoveries, or checks. The required output schema is
\(\langle\textsc{transferable-schema}\rangle\); the quoted input workflows
follow in \(\langle\textsc{source-workflow-blocks}\rangle\). Return only the
consolidated JSON object.
\end{promptbox}

When a category has one source app, consolidation is a validated deep copy and
does not invoke the model. Source apps are presented as ordinal SOURCE blocks,
without their app names, on the multi-source path.

\subsubsection{Target-side reflection}

\begin{promptbox}{Prompt 3: Reward-Grounded Workflow Reflection}
\promptpart{Prompt Template}
You are analyzing an Android GUI agent's target-app rollouts to improve
TRANSFERABLE workflow guidance. The current workflow state for category
\(\langle\textsc{category}\rangle\) is
\(\langle\textsc{current-workflow}\rangle\). The evaluated trajectories are
\(\langle\textsc{rollouts-with-task-seed-reward}\rangle\).

Analyze the supplied trajectories and their rewards. Determine whether an
abstract step is missing or misleading, whether an observed failure or recovery
pattern should be recorded, and whether the completion checks require another
observable or executable signal. Cite trajectory step numbers for every
proposed change.

All suggestions must remain applicable beyond the current interface. Do not
propose app names, screen names, button labels, package/resource identifiers,
coordinates, or literal task-instance values. If the evidence is insufficient,
state that no revision is supported.

\tcbline
\promptpart{Empty-State Branch}
If \(\langle\textsc{current-workflow}\rangle\) is empty, construct an initial
app-agnostic workflow entry from the target evidence instead of editing a
source entry. Use only patterns supported by the supplied target rollouts; an
empty proposal is preferable to an unsupported rule.
\end{promptbox}

\subsubsection{Typed workflow revision}

\begin{promptbox}{Prompt 4: Typed Transferable-Workflow Diff}
\promptpart{Prompt Template}
Given the current transferable workflow state and the grounded reflection,
return a JSON diff. Every operation must target \texttt{layer2/category}; no
app-bound operation is allowed. The only operation types are \texttt{add},
\texttt{modify}, and \texttt{remove}. Include only evidence-supported changes.
For a modified list, return the complete updated list rather than only its new
items. Prefer one to three focused operations and consolidate near duplicates.

Keep at most 16 abstract steps, 20 failure modes, and 16 verification items in
an entry. No app name, package/resource identifier, concrete widget label,
coordinate, state identifier, or literal task-instance value may occur. Return
only the JSON object.

\tcbline
\promptpart{Output Schema}
\texttt{\{"ops": [\{"layer": "layer2", "target": "category",}

\texttt{"op": "add|modify|remove", "key": <category>,}

\texttt{"value": <complete typed value>\}],}

\texttt{"reflection\_summary": <string>, "layer\_tag": "layer2"\}}
\end{promptbox}

For an empty target-grounded state, the first valid diff uses an
\texttt{add/category} operation containing the initial precondition, abstract
steps, failure modes, and verification checklist. It enters the population
without reward and is tested only in later rounds.

\subsubsection{Agent-facing workflow injection}

\begin{promptbox}{Prompt 5: Agent-Facing Workflow Guidance}
\promptpart{Injected Prompt Slot}
\# Workflow knowledge

The following abstract workflow patterns encode general strategies, common
failure modes, and verification signals. They may come from eligible source
workflows, validated target-adaptation experience, or both. They do not
describe this specific app's UI; rely on the current screenshot to identify
actual elements.

\(\langle\textsc{rendered-transferable-workflow-state}\rangle\)
\end{promptbox}

The slot is inserted after the stable agent instruction prefix and before the
episode-specific goal, history, screenshot, and UI-element list. The rendered
state contains the retrieved source entries together with the currently
selected target-grounded revisions. For a Category-Novel App, the initial
rendered state contains no abstract categories; guidance appears after an
empty-state candidate has been constructed, admitted, and selected in a later
round.

\subsection{Validation contracts}

The prompts above state semantic evidence and transfer constraints, while the
runtime applies separate programmatic checks. Table~\ref{tab:validation-contracts}
distinguishes these two roles. This distinction is important: instructions
against concrete UI labels, coordinates, launcher behavior, and literal
task-instance values are semantic prompt constraints; the hard transfer linter
detects identifiable source-app strings, package or resource identifiers, and
app-specific state identifiers. Structural validity and later task reward are
checked independently.

\begin{table}[!ht]
\small
\centering
\setlength{\tabcolsep}{4pt}
\begin{tabularx}{\textwidth}{@{}p{0.16\textwidth}XX@{}}
\hline
\textbf{Stage} & \textbf{Contract} & \textbf{Failure handling} \\
\hline
Source synthesis &
The response must parse as a JSON object containing the typed app-bound and
transferable workflow fields. Model-authored metadata are discarded; the
caller records the evidence count, synthesis model and settings, and a hash of
the complete prompt. &
A malformed response or a missing required component is rejected rather than
converted into a source workflow artifact. \\

Source consolidation &
Entries are merged only within an externally defined app category. The result
must contain at least one typed transferable entry. The hard source-library
linter checks the union of contributing source apps for app-name variants,
package/resource identifiers, and app-specific state identifiers. &
An empty, malformed, or lint-failing library is not written and therefore
cannot be retrieved for a target app. \\

Target reflection &
Only adaptation-task goals, reset identities, compact traces, and executable
task rewards are supplied. The prompt requires every suggested change to cite
adaptation-trace evidence and remain app-agnostic. &
Reflection text cannot directly modify the workflow state; it is used only as
evidence for the subsequent typed-diff request. \\

Typed revision &
The response must parse into typed \texttt{add}, \texttt{modify}, or
\texttt{remove} operations. In target adaptation, only
\texttt{layer2/category} operations are retained, and the applied diff must
change the parent state. &
Malformed JSON is retried up to three times. Empty diffs, all-skipped
operations, app-bound operations, and unchanged children are rejected. \\

Candidate admission &
The caller records the parent, task, iteration, and revision summary. A valid
child enters the population without inheriting the rewards of the rollouts
that produced it. &
The candidate affects final context selection only after it is sampled and
rated using subsequent matched target rollouts. \\

Frozen evaluation &
The selected context and LoRA adapter are loaded as immutable artifacts, and
only the held-out manifest is opened. &
The reflector, population controller, and optimizer are disabled; evaluation
trajectories and rewards cannot alter either adapted state. \\
\hline
\end{tabularx}
\caption{Prompt-level and programmatic validation contracts. Structural
admission determines whether a candidate can be evaluated; executable rewards
from later matched rollouts determine whether it is behaviorally useful.}
\label{tab:validation-contracts}
\end{table}

The validation pipeline therefore does not treat a syntactically valid model
response as learned knowledge. Source entries must satisfy construction and
transfer checks, and target revisions must first become typed candidate states.
Only candidates that subsequently receive favorable executable task feedback
can be selected as the frozen workflow context.